\documentclass{article}
\pdfoutput=1

\usepackage[final]{neurips_2026}
\makeatletter
\renewcommand{\@noticestring}{NeurIPS 2026 Workshop on
Continual Learning for Foundation Models and Agents (CL4FMAgents).}
\makeatother

\usepackage[utf8]{inputenc}
\usepackage[T1]{fontenc}
\usepackage{hyperref}
\usepackage{url}
\usepackage{booktabs}
\usepackage{amsfonts}
\usepackage{amsmath}
\usepackage{amssymb}
\usepackage{amsthm}
\usepackage{nicefrac}
\usepackage{microtype}
\usepackage{graphicx}
\usepackage{xcolor}

\newtheorem{theorem}{Theorem}

\usepackage{listings}
\newcommand{\rankme}{\mathrm{RankMe}}
\newcommand{\rankmec}{\mathrm{RankMe}^{c}}
\newcommand{\reff}{r^{\Sigma}_{\mathrm{eff}}}
\newcommand{\kk}{k_{95}}
\newcommand{\dH}{\Delta H}
\newcommand{\xsim}{s_{\mathrm{cb}}}
\newcommand{\ARL}{\mathrm{ARL}_0}

\title{When Rank Rises as LLMs Degrade}

\author{%
  Zhaohui Geoffrey Wang \\
  University of Southern California \\
  \texttt{zwang000@usc.edu} \\
}

\begin{document}

\maketitle

\begin{abstract}
Post-training adapts a language model in a non-stationary environment, and
practitioners increasingly monitor \emph{representation health} --- RankMe and
related spectral statistics --- to decide when adaptation has gone wrong. Such
monitors carry an implicit directional assumption inherited from the
self-supervised vision literature: rank goes \emph{down} when representations
degrade. We show that this assumption is not safe under LLM post-training. In a
controlled sweep over degradation modes
(Qwen3-0.6B, four modes $\times$ three seeds), a
data-duplication regime that degrades held-out loss by $75\%$ relative to a
healthy run drives RankMe \emph{above} healthy --- in its original uncentred
form as well as the centred variant ($13.5$ pooled s.d.) --- and the
covariance effective rank to nearly twice healthy: damage there is spectral
\emph{dispersion}, not collapse, so a one-sided monitor scores the worst
checkpoint as the healthiest. A learning-rate misconfiguration, by contrast,
moves $\rankmec$ and $\kk$ in the conventional direction while the original
uncentred RankMe is inconsistent across seeds --- so the sign is a property of
the regime--statistic pair, and cannot be fixed by recalibration. We also separate two statistics the literature often conflates:
RankMe normalises singular values whereas covariance effective rank normalises
eigenvalues, and on raw intermediate-layer hidden states the
massive-activation phenomenon pins the latter near $1$ out of $d$ on an
\emph{unmodified pretrained checkpoint} while leaving the former with usable
range. We then stress-test the natural
repair --- \emph{two-sided}, multi-channel sequential monitoring with a
multiplicity-aware false-alarm target --- with calibration and test data held
strictly apart.
In a pre-registered shared-prefix, leave-one-seed-out evaluation on
Qwen3-0.6B, the two-sided ensemble detects all three damage regimes in every
fold within 10--60 steps of the fork, and the firing direction separates
spectral dispersion from downward-rank damage; but it leads the
held-out probe loss only once in nine folds (one 10-step interval), and
with two
calibration seeds it does not achieve zero false alarms on the held-out healthy
seed. We report this negative result in full: spectral monitoring diagnoses the
failure regime, it does not warn consistently earlier than a held-out loss, and
validity claims made without a held-out healthy seed should not be trusted.
\end{abstract}

\section{Introduction}
\label{sec:intro}

Foundation models are no longer trained once. They are updated continually:
post-trained on new instruction data, adapted to a new domain, re-aligned,
patched. Each update is a step in a non-stationary environment, and each step
can damage what the previous ones built. The operational question for anyone
running such a pipeline is not whether damage is possible but whether it can be
\emph{noticed in time} --- ideally from an unlabelled signal that is cheap to
compute at every checkpoint, before the evaluation suite has been run.

The community's default answer to that question is spectral. RankMe
\citep{garrido2023rankme}, the effective rank, and related statistics of the
hidden-state covariance are cheap, label-free, and come with a clean intuition
inherited from self-supervised vision: representations degrade by
\emph{collapsing}, so rank goes down, so a falling rank is an alarm. This
paper asks whether that intuition survives contact with LLM continual
post-training, and whether the resulting monitor can be made reliable enough
to act on. The narrow version of the question is: \emph{can representation
statistics serve as a reliable, label-free monitor of continual updates?}

Our answer is a qualified no, established in three steps. First, an exact
metric audit: RankMe and the covariance effective rank are different
statistics (singular values versus eigenvalues), and on raw LLM hidden states
massive activations \citep{sun2024massive} pin the latter near $1$ of $d$ at
intermediate layers of an unmodified checkpoint, so any dynamic range reported
for it there without per-token normalisation is an artefact. Second, a regime-dependent sign
inversion: under data-duplication overfitting the held-out loss degrades by
$75\%$ while every form of RankMe and the effective rank \emph{rise} far
above healthy, whereas a learning-rate misconfiguration moves the centred
statistics conventionally and the uncentred original inconsistently. The sign
is a property of the regime--statistic pair; even centring can change the
direction a monitor observes. Third, and this is
where the stress test bites, we evaluate the obvious repair --- a two-sided,
multi-channel sequential test with a union-bound false-alarm correction ---
under a pre-registered protocol in which every branch forks from a shared
healthy checkpoint and every threshold is calibrated on seeds the detector
never sees. The ensemble detects all three damage regimes in every fold, and
the direction that fires separates dispersion-type damage from the two
downward-rank regimes. But its lead over the
held-out probe loss is at most one probe interval, in one of nine folds, and
with two calibration seeds it does not achieve zero
false alarms on a held-out healthy seed.

The negative result is the point. A monitor anti-correlated with damage is
worse than none, because it licenses continued training; a monitor that
fires with the loss it is meant to anticipate adds diagnosis, not warning;
a false-alarm guarantee certified only on calibration data is not a
guarantee. We report all three so the next monitor is evaluated against
them.

\paragraph{Contributions.}
\begin{enumerate}
  \item \textbf{An exact metric audit.} RankMe (singular values) and the
  covariance effective rank (eigenvalues) are computed under their original
  definitions and shown not to be interchangeable on LLM hidden states;
  massive activations remove the dynamic range of the latter on raw
  intermediate-layer states unless tokens are normalised
  (\S\ref{sec:instrument}).
  \item \textbf{A regime--statistic sign structure.} Under data-duplication
  overfitting every rank statistic rises well above healthy while held-out
  loss degrades sharply; under learning-rate misconfiguration the centred
  statistics fall while uncentred RankMe is inconsistent across seeds. The
  sign is a property of the regime--statistic pair
  (\S\ref{sec:inversion}, \S\ref{sec:instrument}), so a one-sided monitor
  cannot be repaired by recalibration.
  \item \textbf{A fork-valid sequential evaluation.} Detectors are scored on a
  shared-prefix change-point design with leave-one-seed-out calibration, a
  pre-registered loss-only severity gate, and a union-bound multiplicity
  correction for the ensemble stopping time; false alarms are measured on a
  held-out healthy seed (\S\ref{sec:monitor}).
  \item \textbf{A negative result, reported in full.} Under that protocol the
  two-sided ensemble detects every regime and identifies it by direction, but
  provides no consistent early warning over a held-out probe loss and does not achieve
  zero held-out false alarms with two calibration seeds. Spectral signals
  diagnose failure geometry; they do not presently justify autonomous stopping
  (\S\ref{sec:monitor}).
\end{enumerate}

\section{Related Work}
\label{sec:related}

\paragraph{Spectral representation-health metrics.}
RankMe \citep{garrido2023rankme} validated the singular-value entropy as a
label-free proxy for downstream linear-probe accuracy in vision SSL; the
variance term of VICReg \citep{bardes2022vicreg} and the uniformity loss of
\citet{wang2020understanding} encode the same conviction that healthy
representations occupy many directions, degraded ones few. All of these were
validated in a setting whose canonical failure is rank \emph{loss}, and the
same contraction-signed reading runs through the loss-of-plasticity
literature, where falling feature or effective rank accompanies lost
adaptability \citep{kumar2021implicit,lyle2023understanding,dohare2024plasticity}.
We test whether the machinery transfers to LLM continual post-training and
find that the statistics remain computable but their directional semantics do
not transfer: the sign of the response depends on the failure regime
(\S\ref{sec:inversion}). The anisotropy literature's rogue dimensions
\citep{ethayarajh2019contextual,timkey2021bark} are the massive-activation
geometry \citep{sun2024massive} that floors the covariance effective rank in
\S\ref{sec:instrument}.

\paragraph{Collapse taxonomies.}
\citet{kim2025taxonomy} separate complete from dimensional collapse and
catalogue detectors for each; both classes contract the spectrum. Our
contribution is not a new taxonomy but the observation that LLM post-training
damage can present as a third geometry --- \emph{dispersion}, a spectrum
spreading out while held-out NLL degrades --- which contraction-oriented
taxonomies do not cover and contraction-oriented detectors mis-sign.

\paragraph{Online training monitors.}
The Collapse Index \citep{collapseindex2026} is the closest work: an online,
topology-derived early-warning statistic for representation collapse,
evaluated on LLM fine-tuning among other settings. It monitors a different
signal (a Morse-theoretic summary rather than the spectrum), and its
evaluation reports the statistic's trajectory rather than a calibrated
stopping rule. Our contribution is complementary: a protocol --- shared-prefix
forks, leave-one-seed-out calibration, held-out healthy false alarms, lead
time against a frozen loss breach --- that any early-warning-type claim,
theirs included, needs in order to be certified. Under that protocol our own
spectral ensemble does not certify (\S\ref{sec:monitor}).

\paragraph{Sequential change detection.}
Our detector is Page's CUSUM \citep{page1954continuous} applied per channel
and direction, with the ensemble stopping time controlled by the
fixed-horizon union bound of \eqref{eq:union}; nothing here is novel as
statistics, deliberately, so that what is being evaluated is the signal, not
the test. Anytime-valid alternatives (e-processes, confidence sequences) would make
the bound uniform in time, but inherit the same load-bearing premise our
data fails to certify: a correctly specified healthy null.

\section{Setup}
\label{sec:setup}

\subsection{Diagnostics}
Let $Z \in \mathbb{R}^{N \times d}$ collect hidden-state representations of a
\emph{fixed} held-out probe set, with singular values
$\sigma_1 \ge \cdots \ge \sigma_d \ge 0$, and let $\Sigma$ be the covariance
of its rows (i.e.\ of the \emph{centred} matrix $Z-\bar Z$) with eigenvalues
$\lambda_1 \ge \cdots \ge \lambda_d > 0$ and trace $T$. We track:

\begin{align}
\rankme &= \exp H(p), \quad p_k = \sigma_k / \textstyle\sum_j \sigma_j + \epsilon
  &&\text{(RankMe, \citealp{garrido2023rankme})} \label{eq:rankme}\\
\reff &= \exp\Big(-\textstyle\sum_i q_i \log q_i\Big), \quad q_i = \lambda_i / T
  &&\text{(effective rank)} \label{eq:reff}\\
\kk &= \min\{k : \textstyle\sum_{i\le k}\lambda_i \ge 0.95\,T\}
  &&\text{(95\%-energy rank)} \label{eq:k95}\\
\dH &= \tfrac{d}{2}\log(T/d) - \tfrac12 \log\det\Sigma
  &&\text{(anisotropy deficit)} \label{eq:dH}\\
\xsim &= \textstyle\frac{1}{N(N-1)}\sum_{i \ne j}
  \frac{\langle z_i, z_j\rangle}{\|z_i\|\|z_j\|}
  &&\text{(cross-batch similarity)} \label{eq:xsim}
\end{align}
$\rankme$ follows its original definition: the SVD is taken on $Z$ as the
model emits it, \emph{without centring}, with $\epsilon=10^{-7}$. We write
$\rankmec$ for the centred variant that reconstructs
$\sigma_k \propto \sqrt{\lambda_k}$ from the covariance spectrum; the two
differ by the rank-one mean update $Z^{\top}Z/N = \Sigma\,(N{-}1)/N + \bar z
\bar z^{\top}$ and coincide only when $\bar z = 0$. We report both wherever
they differ materially. One dependency is exact: under per-token RMS
normalisation ($\|z_i\|^2 = d$, \S\ref{sec:probe}) and the $1/(N{-}1)$
convention, $\xsim = 1 - \operatorname{tr}\Sigma/d$, so $\xsim$ and the
covariance trace are one degree of freedom, not independent channels; the
separate evidence for \emph{dispersion} is the centred spectral shape.
Computation details: Appendix~\ref{app:impl}.

Held-out NLL on the same fixed probe set is our \emph{pre-registered offline
damage proxy}: the operational damage criterion frozen by our protocol, against
which every detector is scored. It is a proxy --- downstream capability is not
measured directly (\S\ref{sec:limits}) --- but it is the strongest cheap
baseline available, which is exactly what makes it the right bar.

\subsection{Degradation modes}
\label{sec:modes}
All runs post-train Qwen3-0.6B \citep{qwen3} (hidden size $d=1024$, 28
layers) with full fine-tuning, AdamW, batch $4\times512$ tokens, base learning
rate $10^{-5}$, on databricks-dolly-15k unless stated. The probe is a fixed
256-sequence tail of the same corpus, disjoint from every training set.
\begin{center}\small
\begin{tabular}{@{}llp{6.6cm}@{}}
\toprule
mode & role & recipe \\
\midrule
healthy & reference & Dolly, $1\times$ lr, warm-up 20 \\
domain\_shift\_v2 & \emph{venue-fit scenario} & continual post-training on
  GSM8K \citep{cobbe2021gsm8k}, probe unchanged; $8\times$ lr from a
  pre-specified loss-only gate on seed~0 (App.~\ref{app:severity}) \\
high\_lr & controlled failure mechanism & Dolly, $15\times$ lr, no warm-up
  (calibrated, App.~\ref{app:severity}) \\
duplicate\_data & controlled failure mechanism & 8 unique Dolly examples
  cycled, $1\times$ lr \\
narrow\_domain (v1) & negative control & Dolly \texttt{closed\_qa} only,
  $1\times$ lr; probe loss $+0.06$ nats --- damage gate never triggered \\
\bottomrule
\end{tabular}
\end{center}
Phase 1 (\S\ref{sec:inversion}) runs each mode from step $0$ for 1000 steps,
three seeds, probed every 25 steps; it supports seed-replicated endpoint
comparisons only. Phase 2 (\S\ref{sec:monitor}) forks every mode from a shared
300-step healthy prefix per seed.

\subsection{Probe protocol}
\label{sec:probe}
The probe is a fixed set of 256 held-out Dolly sequences (the tail of a
seed-0 shuffle, disjoint from every training corpus), tokenized once at length
256 and reused verbatim at every measurement, so any movement in the logged
signals is the model changing, not the data. Rows of $Z$ are \emph{token}
representations: every non-pad position of the probed layer contributes one
row ($N\approx31$k $\gg d=1024$); padding is excluded by the attention mask,
and pad labels are masked to $-100$ in the NLL. Unless stated, the layer is
the last hidden layer, taken \emph{after} the model's final RMSNorm
(Appendix~\ref{app:impl}), and each token vector is RMS-normalised
(\S\ref{sec:instrument}) before any second-moment statistic --- a diagnostic
choice distinct from the model's own norm. $\Sigma$ is
accumulated in float64 (streaming Welford) from bfloat16 activations;
$\rankme$ uses the uncentred second moment $Z^\top Z$, also in float64, with
$\epsilon=10^{-7}$ in \eqref{eq:rankme}. $\dH$ requires $\Sigma$ to actually
be full-rank, for which $N>d$ is only necessary; we verify the numerical rank
at every measurement and flag $\dH$ invalid otherwise (it never triggered on
token-level data; sequence-pooled data has $N<d$ and $\dH$ is not reported
there).

\section{Two statistics, not one, and a normalisation prerequisite}
\label{sec:instrument}

\paragraph{RankMe is not the covariance effective rank.}
$\rankme$ \eqref{eq:rankme} is an entropy over normalised \emph{singular
values} of the uncentred $Z$; $\reff$ \eqref{eq:reff} is an entropy over
normalised covariance \emph{eigenvalues}. They differ twice over: by a square
(for centred data $\lambda_k \propto \sigma_k^2$, so $\rankmec$ weights
$\sqrt{\lambda_k}$ and is far less sensitive to a single dominant direction)
and by centring (the mean adds a rank-one term to the uncentred second moment,
and LLM hidden states have large means). Substituting one for the other, as
covariance-based diagnostic code does silently, changes the statistic. We
report them separately throughout.

\paragraph{The distinction is not cosmetic.}
Table~\ref{tab:massive} reports an unmodified pretrained checkpoint. At an
intermediate layer a single direction carries $99.5\%$ of the trace --- the
massive-activation phenomenon \citep{sun2024massive}, not collapse. Under that
spectrum $\reff = 1.06$ out of $d = 1024$: the covariance effective rank sits at
its floor \emph{on a healthy model}, with no room left to fall, so no threshold
can make it a detector. $\rankme$ on the same representations is $120.13$
(centred variant $\rankmec = 121.24$), two orders of magnitude higher, and
retains usable range. \textbf{The flooring claim holds for the covariance
effective rank and not for RankMe, in either the original uncentred or the
centred form.}

\begin{table}[t]
\caption{Effect of massive activations, and of per-token normalisation, on the
statistics. Qwen3-0.6B, $d{=}1024$; the baseline here is the \emph{unmodified
pretrained checkpoint} (no fine-tuning, not a post-training endpoint),
token-level; $\rankme$ is the original uncentred definition,
$\rankmec$ its centred variant. The covariance effective rank is floored on
raw intermediate-layer states; neither form of RankMe is.}
\label{tab:massive}
\centering
\begin{tabular}{llrrrrr}
\toprule
transform & layer & $\reff$ & $\reff/d$ & $\rankme$ & $\rankmec$ & $\lambda_1/T$ \\
\midrule
none    & 14 (of 28) & \textbf{1.06} & \textbf{0.001} & 120.13 & 121.24 & 0.995 \\
none    & last       & 192.90 & 0.188 & 770.63 & 822.50 & 0.134 \\
RMS     & 14 (of 28) & 308.96 & 0.302 & 718.24 & 782.25 & 0.058 \\
RMS     & last       & 192.19 & 0.188 & 765.11 & 818.28 & 0.131 \\
\bottomrule
\end{tabular}
\end{table}

Per-token RMS normalisation restores the covariance effective rank at the
intermediate layer ($1.06 \to 308.96$). RMS and $\ell_2$ normalisation give
numerically identical diagnostics, as they must: they differ by the constant
factor $\sqrt{d}$, and every quantity in \eqref{eq:reff}--\eqref{eq:xsim} is
scale-invariant. We adopt per-token RMS normalisation with covariance (not
correlation) scaling throughout, since correlation scaling divides out the
per-dimension variance disparity that is itself part of the signal.

\paragraph{Centring is part of the definition.}
The original $\rankme$ and its centred reconstruction $\rankmec$ can also
part company with \emph{each other}. On shared-prefix branches
(\S\ref{sec:monitor}), under duplicate\_data both rise far above the healthy
drift ($\rankme$: $+96$ to $+109$ against healthy $+2$), but under high\_lr
$\rankmec$ falls in all three seeds ($-4.7$ to $-16.4$) while $\rankme$
\emph{rises} in two of three ($+14.4$, $+10.5$, $-1.5$). Whether the
representation matrix is centred before the SVD --- a detail implementations
rarely report --- changes not just the value but, in some regimes, the
direction a monitor sees (per-seed values in Appendix~\ref{app:traj}).

\section{Rank metrics can invert under post-training}
\label{sec:results}
\label{sec:inversion}

\paragraph{Unit of inference.}
Each (mode, seed) run is one trajectory, and a trajectory is one observation.
Checkpoints within a run are strongly autocorrelated, so pooling them would
inflate the apparent sample size; all figures below are endpoint values
aggregated \emph{across seeds}, with $n$ equal to the number of runs. We report
separation in pooled standard deviations and refuse to report an effect size
where $n=1$.

\begin{table}[t]
\caption{Endpoint diagnostics after $1000$ steps, mean $\pm$ s.d.\ over $3$
seeds; separation from healthy in pooled s.d.\ in brackets, defined as
$|\mu_m-\mu_h| / \sqrt{\sigma_m^2+\sigma_h^2}$ over seed endpoints ($m$ =
mode, $h$ = healthy). Held-out probe loss
is the pre-registered offline damage proxy. Arrows give the direction relative to
healthy; \textbf{bold} marks a move opposite to the conventional
``rank falls under degradation'' reading.}
\label{tab:main}
\centering
\footnotesize
\setlength{\tabcolsep}{2.5pt}
\begin{tabular}{@{}lrrrrr@{}}
\toprule
mode & probe loss & $\rankmec$ & $\reff$ & $\kk$ & $\xsim$ \\
\midrule
healthy & $2.530\pm0.01$ & $812.6\pm0.4$ & $174.8\pm1.6$ & $689.7\pm0.6$ & $0.394\pm0.00$ \\
\midrule
high\_lr & $3.153\pm0.16$
  & $799.9\pm3.5$ $\downarrow$\,{\tiny[3.6]}
  & $183.2\pm4.2$ {\tiny[1.9, n.s.]}
  & $665.7\pm5.9$ $\downarrow$\,{\tiny[4.1]}
  & $0.277\pm0.01$ $\downarrow$\,{\tiny[7.8]} \\
duplicate\_data & $\mathbf{4.429\pm0.31}$
  & $\mathbf{877.9\pm4.8}$ $\uparrow$\,{\tiny[13.5]}
  & $\mathbf{329.2\pm14.4}$ $\uparrow$\,{\tiny[10.7]}
  & $\mathbf{782.0\pm7.0}$ $\uparrow$\,{\tiny[13.1]}
  & $0.197\pm0.01$ $\downarrow$\,{\tiny[17.6]} \\
narrow\_domain & $2.586\pm0.01$
  & $\mathbf{816.4\pm0.8}$ $\uparrow$\,{\tiny[4.2]}
  & $179.4\pm1.1$ {\tiny[2.4]}
  & $\mathbf{696.3\pm1.5}$ $\uparrow$\,{\tiny[4.1]}
  & $0.368\pm0.00$ $\downarrow$\,{\tiny[7.5]} \\
\bottomrule
\end{tabular}
\end{table}

\paragraph{The inversion is regime-dependent, and it replicates.}
Under \texttt{duplicate\_data}, held-out loss degrades by $75\%$ relative to
healthy, yet $\rankmec$ rises above healthy ($877.9$ vs $812.6$, $13.5$ s.d.) and
$\reff$ nearly doubles ($329.2$ vs $174.8$, $10.7$ s.d.): by the conventional
reading the worst model has the richest representations, in all three seeds.
Under \texttt{high\_lr}, also genuinely damaged (loss $3.153$ vs $2.530$),
$\rankmec$ and $\kk$ move the \emph{other} way ($3.6$ and $4.1$ s.d.\ down)
--- while on the shared-prefix branches the original uncentred $\rankme$ is
inconsistent across seeds under the same regime (\S\ref{sec:instrument}).
\textbf{The sign is a property of the regime--statistic pair; even centring
can change the direction a monitor observes.} A monitor that fixes a
direction in advance is mis-specified for at least one such pair, and
recalibration cannot repair that.

\paragraph{What we do not claim.}
$\reff$ also rises under \texttt{high\_lr}, but by $1.9$ s.d., inside seed
noise; we do not count that as an inversion. \texttt{narrow\_domain} degrades
held-out loss by only $2.2\%$, which we regard as too small to call damage even
though its metric shifts are statistically separable; it is reported as a
negative result for the mode, not as a third inversion. The inversion claim
therefore rests on \emph{one} regime replicated across three seeds, not on the
sweep as a whole.

\paragraph{Mechanism: dispersion, not collapse.}
This is not numerical failure. The numerical rank of $\Sigma$ remains full
throughout, and every quantity plateaus by step $200$ and holds to step $1000$.
What moves is where the variance sits: $\lambda_1$ \emph{falls} from $84.79$ to
$74.53$ while $T$ \emph{rises} from $649.6$ to $820.3$. Variance is
redistributed out of the dominant direction into the tail. In this regime overfitting is
\emph{accompanied by} spectral dispersion on the held-out probe --- we observe
the redistribution, not its mechanism. Damage here is dispersion, not
collapse.

\paragraph{Why not just monitor the loss?}
Held-out NLL is an offline damage proxy, not a deployment-time oracle in every
setting --- but for a language model it does not require labels: on a fixed
text stream the next token is its own supervision, and our probe is plain
held-out text. It is therefore self-supervised, inexpensive, and a
deliberately strong baseline. The stricter question a spectral monitor must
answer is whether it \emph{warns before} this baseline breaches a frozen
threshold; Table~\ref{tab:detect} answers it in the negative. Spectral
statistics retain two potential advantages the NLL lacks --- they need no
probe forward passes beyond the ones already made, and they expose the
\emph{geometry} of the failure --- and the second survives our stress test
while the first does not translate into earlier warning.

\paragraph{A channel with a consistent sign, with its direction flipped.}
$\xsim$ falls below healthy in \emph{every} mode we ran --- $7.8$ s.d.\ under
\texttt{high\_lr}, $17.6$ s.d.\ under \texttt{duplicate\_data}, $7.5$ s.d.\
under \texttt{narrow\_domain} --- making it the only channel here whose
direction is consistent across regimes. It is precisely the channel that plays a \emph{secondary} role in the
vision collapse literature, where uniformity collapse makes representations more
\emph{similar}; under LLM post-training the sign reverses, and damage spreads
representations apart. Any monitor ported from vision must re-derive its channel
directions rather than inherit them.
Consistency of sign is all this buys: in the sequential evaluation of
\S\ref{sec:monitor}, $\xsim$ neither leads the loss breach nor passes the
held-out false-alarm bar, so it is a candidate direction convention, not a
working monitor. Its sign is consistent across all \emph{tested update
branches}, including the below-gate narrow\_domain control; whether that
extends to untested damage modes is open.

\section{A two-sided monitor under held-out false-alarm evaluation}
\label{sec:monitor}

The evidence in \S\ref{sec:inversion} dictates the design. Rank channels must
be evaluated in \emph{both} directions, because both inflation and collapse
indicate damage --- and damaged branches never participate in direction or
threshold selection. Because the healthy baseline itself drifts
(\S\ref{sec:setup}), a fixed threshold cannot separate drift from damage;
what is needed is an evidence-accumulating sequential test with an explicit
false-alarm target.

\subsection{The anisotropy-deficit channel}

\begin{theorem}[Anisotropy deficit is non-negative]
\label{thm:dH}
Let $\lambda_1,\dots,\lambda_d > 0$ with $T = \sum_i \lambda_i$. Then
$\dH = \tfrac{d}{2}\log(T/d) - \tfrac12\sum_i \log\lambda_i \ge 0$,
with equality if the spectrum is flat.
\end{theorem}

\noindent
The converse (equality \emph{only} if the spectrum is flat) also holds, by
strict concavity of $\log$, but we state the theorem in the direction we have
machine-checked; see Appendix~\ref{app:lean}.

$\dH$ is the gap between the differential entropy of the isotropic Gaussian with
the same total variance and that of $\mathcal{N}(0,\Sigma)$; the bound is the
arithmetic--geometric mean inequality on the spectrum, machine-checked in
Lean~4 against Mathlib with an axiom audit confirming no appeal to
\texttt{sorry} (Appendix~\ref{app:lean}). We note it only because $\dH$ moves in
the unexpected direction in Table~\ref{tab:main}, so its floor is one property
that should not rest on inspection.

\subsection{Evaluation design: shared prefix, held-out calibration}
\label{sec:design}

Two easy design errors each manufacture success.

\paragraph{Calibrating and testing on the same runs.}
Choosing the smallest margin that gives zero false alarms on the healthy runs
and then reporting zero false alarms on those same runs is a definition, not a
result. We instead calibrate margins, channel directions and channel selection
on a strict subset of healthy seeds, evaluate the false-alarm rate on the
held-out healthy seed, and rotate over folds. Damaged trajectories never
participate in calibration.

\paragraph{No common pre-change history.}
A sequential change detector is meant to locate the moment a stream's
distribution changes. Runs that begin under different recipes at step $0$ have
no change point to find, which reduces the task to separating two populations.
We therefore adopt a \emph{shared-prefix} design: each seed trains an identical
healthy prefix, forks from that common checkpoint into each mode at a known
switch point, and detectors are scored by \emph{detection delay} after the fork
and by lead time relative to the held-out-loss breach.
Concretely (protocol frozen before any damaged branch was inspected;
\texttt{DETECTOR\_PROTOCOL.md}, supplement): three seeds each train a healthy
$\tau=300$-step prefix; its checkpoint (weights and optimiser state, SHA-256
hashed) forks into healthy, high\_lr, duplicate\_data and domain\_shift\_v2
branches of $600$ steps, probed every $10$ steps. For each
held-out seed $s$, baseline $\mu_c$, scale $\sigma_c$ and threshold $h_c$ of
every channel are frozen from the \emph{healthy branches of the other two
seeds only}; the loss-breach level $L=\mu_L+3\sigma_L$ is frozen the same way.
Every channel is tested in both directions with slack $k=0.5$; directions are
not chosen from damaged data. False alarms are scored on the held-out healthy
branch, delay on the held-out damaged branches.

\subsection{Sequential test and a multiplicity-correct guarantee}
\label{sec:multiplicity}
The ensemble monitors the five channels of \S\ref{sec:setup}
($\rankmec,\reff,\kk,\dH,\xsim$) in both directions ($C{=}5$, $2C{=}10$
one-sided tests); each runs a CUSUM $S^{\pm}_t=\max(0,S^{\pm}_{t-1}\pm z_t-k)$
on the standardised series and alarms at $S^{\pm}_t>h_c$.
Monitoring $C$ channels in both directions and stopping at the first alarm makes
the stopping time $\tau_{\mathrm{ens}}=\min_{c,\pm}\tau_{c,\pm}$ a minimum over
$2C$ dependent stopping times, so a per-channel false-alarm rate does not
transfer: the ensemble $\ARL$ is strictly worse than that of any single
channel. The guarantee we target is the fixed-horizon union bound: if under
the healthy distribution $\Pr_0(\tau_{c,\pm}\le H)\le\alpha/(2C)$ for every
one-sided test, then without any independence assumption
\begin{equation}
\Pr_0(\tau_{\mathrm{ens}}\le H)\;\le\;\sum_{c,\pm}\Pr_0(\tau_{c,\pm}\le H)\;\le\;\alpha,
\qquad\text{hence}\qquad
\ARL=\mathbb{E}_0[\tau_{\mathrm{ens}}]\;\ge\;H(1-\alpha).
\label{eq:union}
\end{equation}
The premise is the load-bearing part, and two calibration seeds cannot certify
it: each one-sided test sees two healthy null series, so its achievable
empirical levels at branch granularity are $\{0,\tfrac12,1\}$, and
$\alpha/(2C)=0.005$ lies below the smallest nonzero level --- the calibrated
threshold degenerates to the zero-crossing threshold (pooling the four
series--direction pairs, as implemented, degenerates identically). The
held-out healthy false alarms in Table~\ref{tab:detect} show the premise is
\emph{not} satisfied at this data scale: \eqref{eq:union} is the guarantee we
target and empirically fail to certify.
A synthetic check (supplement): with $C=12$ i.i.d.\ channels and two
calibration series, zero-calibration-FP thresholds yield held-out per-channel
FP rates of 4--58\% and an ensemble rate of 98.5\%.

\subsection{Detection results}
\begin{figure}[t]
\centering
\includegraphics[width=0.78\linewidth]{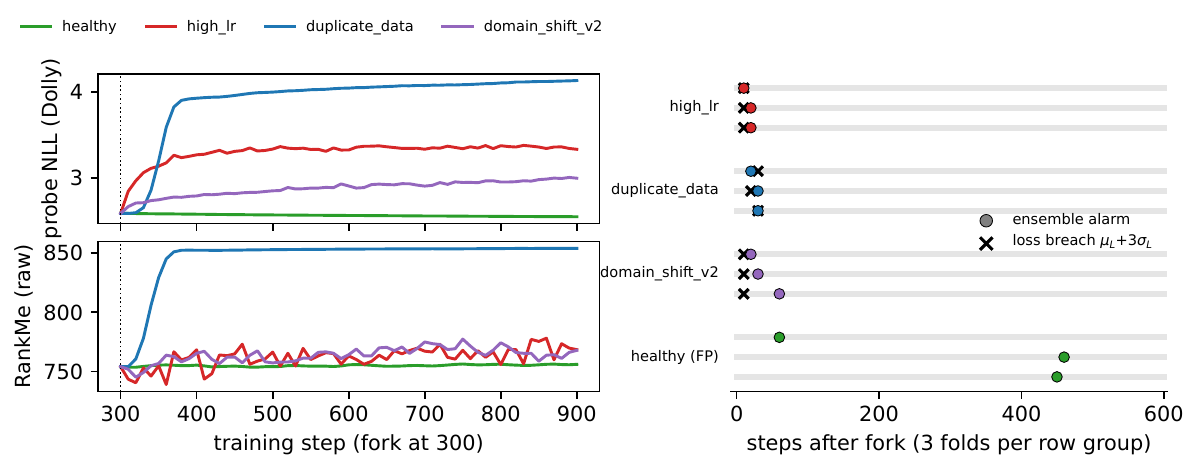}
\caption{The negative result at a glance (Qwen3-0.6B, shared-prefix
branches). \emph{Left} (seed 0): duplicate\_data degrades held-out NLL most
while its raw RankMe rises $\sim$100 above healthy --- dispersion, not
collapse. \emph{Right} (all folds): loss breach ($\times$) and ensemble
alarm ($\circ$) land within 10--60 steps of the fork and of each other ---
no consistent early warning --- while the held-out healthy branch false-alarms
at 60/460/450 steps (folds 0--2, as in Table~\ref{tab:detect}).}
\label{fig:core}
\end{figure}
\begin{table}[t]
\caption{Shared-prefix, leave-one-seed-out detection (Qwen3-0.6B, last
layer). Entries: steps after the fork, range over folds; ``cens.''\,= no alarm
in 600 steps; lead $=$ breach $-$ alarm step per fold (negative $=$ alarm
lags); last column: false alarms on the held-out \emph{healthy} branch.
Top: pre-registered; bottom: post-hoc fork-anchored, no pre-registration
credit. RankMe rows: the protocol-frozen $\rankmec$ (uncentred:
App.~\ref{app:traj}).}
\label{tab:detect}
\centering\footnotesize
\begin{tabular}{lcccc}
\toprule
 & high\_lr & duplicate\_data & domain\_shift\_v2 & healthy (held-out) \\
\midrule
\multicolumn{5}{l}{\emph{pre-registered protocol (constant baseline)}}\\
ensemble alarm & 10--20 & 20--30 & 20--60 & FP@60/460/450 \\
$\rankmec\downarrow$ only & 10--20 & cens. & 20--30 (2/3) & FP@--/530/-- \\
$\rankmec\uparrow$ & 550 (1/3) & 20--30 & 120--380 & FP@60/--/-- \\
$\xsim\downarrow$ & 30--40 & 40 & 50--60 & FP@--/--/450 \\
loss breach $\mu_L{+}3\sigma_L$ & 10 & 20--30 & 10 & none \\
lead (per fold) & 0/$-$10/$-$10 & $+$10/$-$10/0 & $-$10/$-$20/$-$50 & -- \\
detected / false-alarm folds & 3/3 & 3/3 & 3/3 & \textbf{3/3} \\
\midrule
\multicolumn{5}{l}{\emph{post hoc: fork-anchored $x_t-x_\tau$}}\\
ensemble alarm & 10--20 & 20--30 & 20--60 & FP@--/140/240 \\
$\rankmec\downarrow$ only & 10--20 & cens. & 20 (2/3) & none \\
$\rankmec\uparrow$ & 30 (1/3) & 20--30 & 60--350 & FP@--/140/-- \\
$\xsim\downarrow$ & 30--40 & 30--40 & 40--80 & FP@--/500/-- \\
lead (per fold) & 0/$-$10/0 & 0/0/0 & $-$10/$-$10/$-$50 & -- \\
detected / false-alarm folds & 3/3 & 3/3 & 3/3 & \textbf{2/3} \\
\bottomrule
\end{tabular}
\end{table}

Figure~\ref{fig:core} and Table~\ref{tab:detect} answer the title's
question, negatively on two of three counts.
\textbf{Detection:} the two-sided ensemble alarms in every fold of every
mode within 10--60 steps, and the firing direction is regime-specific as
\S\ref{sec:inversion} predicted: duplicate\_data is caught by
$\rankmec\uparrow$ while $\rankmec\downarrow$ stays censored --- the
conventional one-sided monitor misses it entirely; high\_lr and
domain\_shift\_v2 fire $\rankmec\downarrow$ first.
\textbf{Lead: none that matters.} The fixed $31$k-token probe makes NLL nearly
noiseless ($\sigma_L\approx0.01$), so $\mu_L+3\sigma_L$ is breached at the
first or second post-fork measurement in every mode, the gradual one included;
the ensemble precedes it in exactly one fold (duplicate\_data, by a single
10-step probe interval; Table~\ref{tab:detect}), and in no other fold does any
channel precede it. \emph{If a held-out loss is available, it
is the faster monitor}; the spectrum adds a diagnosis of the regime, not a
usable earlier warning.
\textbf{False alarms:} the pre-registered protocol false-alarms on the
held-out healthy branch in all three folds despite $12$--$20\sigma$
thresholds: healthy seeds are offset from one another by more than their
within-seed noise ($\rankmec$: $814.2$ vs $812.9/813.1$, $\sigma=0.5$).
Fork-anchoring clears fold~0; folds 1--2 still alarm at 140/240 steps ---
two calibration seeds do not bound a third seed's healthy excursion. At
$n=3$ seeds the supportable statement is not zero false alarms but
time-to-false-alarm $\geq140$ steps vs.\ detection delay $\leq60$.

\paragraph{domain\_shift\_v2 is adaptation, not just harm.}
On held-out GSM8K test NLL (measured throughout, never trained on) the
domain\_shift\_v2 branches improve $1.40\to1.10\pm0.01$ while Dolly retention
degrades $2.58\to2.98\pm0.01$ --- a genuine stability--plasticity trade-off;
healthy branches leave GSM8K flat ($1.38$) and the failure mechanisms damage
both domains (GSM8K $2.25$, $2.63$). The monitoring task is thus the hard one:
flagging the stability cost of an update \emph{succeeding} at its own objective.

\section{Limitations}
\label{sec:limits}

\textbf{Scale.} Every number is from Qwen3-0.6B ($d=1024$); the metric audit
is definitional, but whether the regime-specific directions and the no-lead
result hold at larger scale is untested.
\textbf{Regimes.} The upward inversion rests on one mechanism
(duplicate\_data, three seeds); under high\_lr the centred statistics fall
while uncentred $\rankme$ is seed-inconsistent; narrow\_domain (v1) never
triggered its damage gate.
\textbf{Data and proxy.} The \S\ref{sec:inversion} comparisons fork at step 0
and support seed-replicated endpoint claims only; detection claims rest on the
shared-prefix branches. All diagnostics use one fixed held-out probe whose NLL
stands in for downstream capability (the no-lead bar is same-probe, not
downstream); $\dH$ is reported only at token granularity, where $\Sigma$ is
verifiably full-rank.

\section{Conclusion}
Spectral statistics fail two audits: RankMe and the covariance effective rank
are not interchangeable, and centring alone can flip the direction a monitor
sees. Stress-tested fork-valid, the ensemble detects every regime, leads the
held-out loss only once in nine folds (one interval), and false-alarms on
unseen healthy seeds --- failure-geometry diagnosis, not a licence for
autonomous stopping. Evaluate any such monitor this way, held-out healthy row
included.

\bibliographystyle{plainnat}
\bibliography{references}

\newpage
\appendix

\section{Machine-checked proof of Theorem~\ref{thm:dH}}
\label{app:lean}

Theorem~\ref{thm:dH} ($\Delta H \ge 0$, with equality if the spectrum is
flat) is machine-checked in Lean~4.24.0 against Mathlib. The development
proves \texttt{sum\_log\_le\_mul\_log\_mean} (AM--GM in log form),
\texttt{anisotropy\_deficit\_nonneg}, and
\texttt{anisotropy\_deficit\_eq\_zero\_of\_flat}; note the last is the
``if'' direction only, which is all the paper claims. The axiom audit
(\texttt{\#print axioms}) reports every result closed over exactly
\texttt{propext}, \texttt{Classical.choice}, \texttt{Quot.sound} --- i.e.\
no \texttt{sorry} and no additional axioms. Source
(\texttt{lean/AnisotropyDeficit.lean}, compiled with
\texttt{lean AnisotropyDeficit.lean} under the pinned \texttt{LEAN\_PATH}):
\begin{lstlisting}[basicstyle=\ttfamily\scriptsize,breaklines=true]
/-
Machine-checked proof of the sigma-free anisotropy deficit bound (Theorem 1 in the paper).

For a covariance matrix Sigma with strictly positive eigenvalues lambda_1..lambda_d
and trace T = sum_i lambda_i, define

    DeltaH  :=  (d/2) * log (T/d)  -  (1/2) * sum_i log (lambda_i)
             =  (d/2) * log (T/d)  -  (1/2) * log (det Sigma).

DeltaH is the gap between the differential entropy of the isotropic Gaussian with
the same total variance and that of N(0, Sigma). It is exactly the
arithmetic-mean/geometric-mean gap of the spectrum, hence non-negative, and it
vanishes precisely when the spectrum is flat (Sigma isotropic).

The statement is deliberately phrased over the raw spectrum rather than over
matrices: everything the monitor computes is a function of the eigenvalues, so
this is the form actually used, and it keeps the proof free of any linear-algebra
scaffolding.

Verified with Lean 4.24.0 + Mathlib (see lean/README.md for the exact invocation).
-/
import Mathlib.Analysis.MeanInequalities

open Finset

namespace JepaVerify

variable {d : ℕ}

/-- The uniform weight vector `1/d` on `Fin d` sums to one. -/
private lemma uniform_weights_sum_one (hd : 0 < d) :
    ∑ _i ∈ (univ : Finset (Fin d)), (1 : ℝ) / (d : ℝ) = 1 := by
  have hd' : (d : ℝ) ≠ 0 := Nat.cast_ne_zero.mpr hd.ne'
  rw [Finset.sum_const, Finset.card_univ, Fintype.card_fin, nsmul_eq_mul]
  field_simp

/-- Log-form AM-GM: the sum of logs is at most `d` times the log of the mean.

This is weighted AM-GM with uniform weights, pushed through `log`. -/
theorem sum_log_le_mul_log_mean (hd : 0 < d) (lam : Fin d → ℝ) (hpos : ∀ i, 0 < lam i) :
    ∑ i, Real.log (lam i) ≤ (d : ℝ) * Real.log ((∑ i, lam i) / (d : ℝ)) := by
  have hd0 : (0 : ℝ) < (d : ℝ) := Nat.cast_pos.mpr hd
  have hw : ∀ i ∈ (univ : Finset (Fin d)), (0 : ℝ) ≤ 1 / (d : ℝ) := fun _ _ => by positivity
  have hz : ∀ i ∈ (univ : Finset (Fin d)), (0 : ℝ) ≤ lam i := fun i _ => (hpos i).le
  -- weighted AM-GM with all weights equal to 1/d
  have amgm :=
    Real.geom_mean_le_arith_mean_weighted univ (fun _ => (1 : ℝ) / (d : ℝ)) lam hw
      (uniform_weights_sum_one hd) hz
  have hrhs : ∑ i, (1 : ℝ) / (d : ℝ) * lam i = (∑ i, lam i) / (d : ℝ) := by
    rw [← Finset.mul_sum]; ring
  rw [hrhs] at amgm
  have hprodpos : (0 : ℝ) < ∏ i, (lam i) ^ ((1 : ℝ) / (d : ℝ)) :=
    Finset.prod_pos fun i _ => Real.rpow_pos_of_pos (hpos i) _
  have hlog := Real.log_le_log hprodpos amgm
  rw [Real.log_prod _ _ fun i _ => ne_of_gt (Real.rpow_pos_of_pos (hpos i) _)] at hlog
  have hrw : ∑ i, Real.log ((lam i) ^ ((1 : ℝ) / (d : ℝ)))
      = (1 / (d : ℝ)) * ∑ i, Real.log (lam i) := by
    rw [Finset.mul_sum]
    exact Finset.sum_congr rfl fun i _ => Real.log_rpow (hpos i) _
  rw [hrw] at hlog
  calc ∑ i, Real.log (lam i)
      = (d : ℝ) * ((1 / (d : ℝ)) * ∑ i, Real.log (lam i)) := by field_simp
    _ ≤ (d : ℝ) * Real.log ((∑ i, lam i) / (d : ℝ)) :=
        mul_le_mul_of_nonneg_left hlog hd0.le

/-- **Theorem 1 (anisotropy deficit is non-negative).**

For any strictly positive spectrum, `DeltaH >= 0`. -/
theorem anisotropy_deficit_nonneg (hd : 0 < d) (lam : Fin d → ℝ) (hpos : ∀ i, 0 < lam i) :
    0 ≤ (d : ℝ) / 2 * Real.log ((∑ i, lam i) / (d : ℝ))
          - (1 / 2) * ∑ i, Real.log (lam i) := by
  have h := sum_log_le_mul_log_mean hd lam hpos
  linarith

/-- A flat (isotropic) spectrum attains the bound: `DeltaH = 0`. -/
theorem anisotropy_deficit_eq_zero_of_flat (hd : 0 < d) (c : ℝ) (hc : 0 < c) :
    (d : ℝ) / 2 * Real.log ((∑ _i : Fin d, c) / (d : ℝ))
      - (1 / 2) * ∑ _i : Fin d, Real.log c = 0 := by
  have hd' : (d : ℝ) ≠ 0 := Nat.cast_ne_zero.mpr hd.ne'
  rw [Finset.sum_const, Finset.sum_const, Finset.card_univ, Fintype.card_fin,
      nsmul_eq_mul, nsmul_eq_mul]
  have hcancel : (d : ℝ) * c / (d : ℝ) = c := by field_simp
  rw [hcancel]
  ring

end JepaVerify

/- Audit: each result must rest only on Lean's three standard axioms
   (propext, Classical.choice, Quot.sound). Any appearance of `sorryAx` here
   would mean the proof is incomplete. -/
#print axioms JepaVerify.sum_log_le_mul_log_mean
#print axioms JepaVerify.anisotropy_deficit_nonneg
#print axioms JepaVerify.anisotropy_deficit_eq_zero_of_flat
\end{lstlisting}

\section{Deferred proofs}
\label{app:proofs}
\paragraph{Fixed-horizon union bound (Eq.~\ref{eq:union}).}
Fix a horizon $H$ and let $\tau_{c,\pm}$ be the stopping time of the
one-sided CUSUM on channel $c$ in direction $\pm$ under the healthy
distribution $\Pr_0$. The ensemble stops at
$\tau_{\mathrm{ens}}=\min_{c,\pm}\tau_{c,\pm}$, so
$\{\tau_{\mathrm{ens}}\le H\}=\bigcup_{c,\pm}\{\tau_{c,\pm}\le H\}$ and
subadditivity gives
$\Pr_0(\tau_{\mathrm{ens}}\le H)\le\sum_{c,\pm}\Pr_0(\tau_{c,\pm}\le H)$,
with no independence or stationarity assumption. If each term is at most
$\alpha/(2C)$ the sum is at most $\alpha$. For the $\ARL$ consequence,
Markov's inequality is not needed: since $\tau_{\mathrm{ens}}\ge 0$,
$\mathbb{E}_0[\tau_{\mathrm{ens}}]\ge H\,\Pr_0(\tau_{\mathrm{ens}}>H)\ge
H(1-\alpha)$. The bound is over the fixed horizon only; it says nothing
uniform over time, and it is conditional on the premise
$\Pr_0(\tau_{c,\pm}\le H)\le\alpha/(2C)$, which must itself be certified
from healthy data. With two calibration seeds each one-sided term is
estimated from two null series, giving resolution $1/2$ at branch
granularity, so the premise cannot be certified at $\alpha=0.05$; Table~\ref{tab:detect}'s held-out healthy row
shows it in fact fails at this data scale. Nothing in this appendix is
novel; it is stated to fix exactly what would have been guaranteed had the
premise held.

\section{Implementation details}
\label{app:impl}
Model Qwen3-0.6B \citep{qwen3} ($d=1024$, 28 layers), full fine-tuning in
bfloat16; AdamW, weight decay $0$, gradient clip $1.0$; base learning rate
$10^{-5}$ scaled by the mode's multiplier; linear warm-up over 20 steps where
stated (branches inherit a warm optimiser and use none); batch $4$ sequences
$\times$ $512$ tokens. Probe: 256 sequences $\times$ 256 tokens, batch 8;
layers probed: last and 14; measurement every 25 steps (Phase 1) or 10 steps
(branches). Layer indices address the HF \texttt{output\_hidden\_states}
tuple; its final entry equals \texttt{last\_hidden\_state}, i.e.\ the
last-layer states are taken \emph{after} the model's final RMSNorm (verified
against the transformers Qwen3 forward pass), independently of the diagnostic
per-token normalisation. Spectral accumulators in float64 on-GPU. Fork checkpoints store
model and optimiser state; branches verify the checkpoint SHA-256 against
\texttt{fork.json} before training, and the replay refuses a branch whose
recorded parent hash mismatches. Hardware: one RTX~4090-class 24\,GB GPU per
run; wall-clock $\approx$22\,min per 1000-step Phase-1 run and
$\approx$9--13\,min per 600-step branch, of which $\approx$7\,s per
measurement. Code, frozen protocol (\texttt{DETECTOR\_PROTOCOL.md},
\texttt{DOMAIN\_SHIFT\_V2.md}), unit tests, and all logs are in the
supplement.

\section{Severity calibration}
\label{app:severity}
Both severity knobs were set by pre-specified, loss-only ladders; no
spectral signal was consulted.

\paragraph{high\_lr (Phase 1).} Multipliers $8/15/25\times$ (and $40\times$
in a pilot) for 200 steps: probe NLL ends at $2.78$ / $3.36$ / $4.28$
respectively against a healthy $2.55$; $8\times$ recovers after warm-up,
$25\times$ and $40\times$ collapse within the first 25 steps, leaving no
interval in which a detector could plausibly warn. $15\times$ exits the
healthy envelope while degrading over a resolvable window and was selected.

\paragraph{domain\_shift\_v2 (branches).} Ladder $1/2/4/8\times$ on seed~0
only, gate frozen in \texttt{DOMAIN\_SHIFT\_V2.md}: endpoint NLL at least
$0.30$ nats above the same-seed healthy branch \emph{and} first crossing at
least 50 steps after the fork. Results: $+0.05$ / $+0.06$ / $+0.11$ (fail) /
$+0.44$ with first crossing at 220 steps (pass); seeds 1--2 confirm at
$8\times$ with $+0.45$ (190) and $+0.43$ (220) without re-selection. The
220-step crossing refers to this pre-registered $+0.30$-nat \emph{severity}
gate; Table~\ref{tab:detect}'s 10-step breach uses the much tighter
calibration-defined $\mu_L+3\sigma_L$ \emph{monitoring} threshold --- the
two thresholds answer different questions and do not conflict.

\section{Per-mode trajectories}
\label{app:traj}
Figure~\ref{fig:core} (left) shows the seed-0 trajectories; the table below
gives per-seed shared-prefix branch endpoints (fork at step 300, branch end at
step 900; fork values (seed 0) $\rankme=754.1$, $\rankmec=814.2$,
Dolly NLL $2.58$, GSM8K test NLL $1.40$). $\rankme$ is the uncentred
original, $\rankmec$ the centred variant, both on RMS-normalised token
representations of the last layer.

\begin{center}\small
\setlength{\tabcolsep}{4pt}
\begin{tabular}{@{}llrrrr@{}}
\toprule
mode & seed & $\rankme$ end ($\Delta$) & $\rankmec$ end ($\Delta$) & Dolly NLL & GSM8K NLL \\
\midrule
healthy & 0/1/2 & $756.1/755.2/754.8$ & $813.3/812.7/812.5$ & $2.55/2.53/2.53$ & $1.38/1.38/1.39$ \\
high\_lr & 0 & $768.5$ ($+14.4$) & $809.5$ ($-4.7$) & $3.33$ & $2.22$ \\
 & 1 & $762.8$ ($+10.5$) & $806.2$ ($-6.7$) & $3.14$ & $2.22$ \\
 & 2 & $752.1$ ($-1.5$) & $796.8$ ($-16.4$) & $3.21$ & $2.30$ \\
duplicate\_data & 0 & $853.7$ ($+99.6$) & $880.1$ ($+66.0$) & $4.13$ & $2.48$ \\
 & 1 & $861.8$ ($+109.4$) & $886.9$ ($+74.0$) & $4.60$ & $3.00$ \\
 & 2 & $849.3$ ($+95.8$) & $877.8$ ($+64.6$) & $3.95$ & $2.42$ \\
domain\_shift\_v2 & 0 & $767.8$ ($+13.7$) & $817.4$ ($+3.3$) & $2.99$ & $1.09$ \\
 & 1 & $770.3$ ($+18.0$) & $816.7$ ($+3.8$) & $2.97$ & $1.10$ \\
 & 2 & $769.5$ ($+15.9$) & $816.9$ ($+3.7$) & $2.98$ & $1.11$ \\
\bottomrule
\end{tabular}
\end{center}

Direction summary: duplicate\_data raises both RankMe forms in every seed
(dispersion); high\_lr lowers $\rankmec$ in every seed while $\rankme$ rises
in two of three; domain\_shift\_v2 raises both, mildly and consistently. The
full logs (all channels, every 10 steps, all seeds) are in the supplement.

\end{document}